%% file: MAIN.tex
\documentclass[lettersize,journal]{IEEEtran}
\usepackage{amsmath,amsfonts}
\usepackage{algorithmic}
\usepackage{algorithm}
\usepackage{array}
\usepackage[caption=false,font=normalsize,labelfont=sf,textfont=sf]{subfig}
\usepackage{textcomp}
\usepackage{stfloats}
\usepackage{url}
\usepackage{verbatim}
\usepackage{graphicx}
\usepackage{cite}
\usepackage{xcolor}
\begin{document}

\title{Exploring Transformers for Behavioural Biometrics: A Case Study in Gait Recognition}

%\author{IEEE Publication Technology,~\IEEEmembership{Staff,~IEEE,}
        % <-this % stops a space
%\thanks{This paper was produced by the IEEE Publication Technology Group. They are in Piscataway, NJ.}% <-this % stops a space
%\thanks{Manuscript received April 19, 2021; revised August 16, 2021.}}
\author{
    \IEEEauthorblockN{Paula Delgado-Santos\IEEEauthorrefmark{1}\IEEEauthorrefmark{2}\textsuperscript{\textsection}, Ruben Tolosana\IEEEauthorrefmark{2}\textsuperscript{\textsection}, Richard Guest\IEEEauthorrefmark{1}, Farzin Deravi\IEEEauthorrefmark{1}}, Ruben Vera-Rodriguez\IEEEauthorrefmark{2}
    \\\
    \IEEEauthorblockA{\IEEEauthorrefmark{1}School of Engineering, University of Kent
    \\\{p.delgado-de-santos, r.m.guest, f.deravi\}@kent.ac.uk}
    \\\
    \IEEEauthorblockA{\IEEEauthorrefmark{2}Biometrics and Data Pattern Analytics Lab, Universidad Autonoma de Madrid
    \\\{ruben.tolosana, ruben.vera\}@uam.es}}
%    \\\
%    \IEEEauthorblockA{\IEEEauthorrefmark{1} These authors contributed equally to this research.}

%\markboth{Journal of \LaTeX\ Class Files,~Vol.~14, No.~8, August~2021}{}

\maketitle
\begingroup\renewcommand\thefootnote{\textsection}
\footnotetext{These authors contributed equally to this research.}
\endgroup
\begin{abstract}
%Gait biometric recognition on mobile devices has attracted much attention in recent years, in part, as it is considered to be a user-friendly authentication method. This interest has also been supported by the success of Deep Learning (DL). Architectures based on Convolutional Neural Networks (CNNs) and Recurrent Neural Networks (RNNs) have proven to be convenient for the task, improving performance and robustness of recognition in comparison to traditional machine learning techniques.  This article proposes a novel gait biometric identification systems based on Transformers  for gait biometrics. In addition, new configurations of the Transformers are proposed to further increase their performance. Several state-of-the-art architectures (Vanilla, Informer, Autoformer, Block-Recurrent Transformer, and THAT) are compared with the proposed techniques.  Experiments are carried out using the two popular public databases, whuGAIT and OU-ISIR. The results achieved indicate the potential of the proposed Transformer models to outperform state-of-the-art CNN and RNN architectures.
Biometrics on mobile devices has attracted a lot of attention in recent years as it is considered a user-friendly authentication method. This interest has also been motivated by the success of Deep Learning (DL). Architectures based on Convolutional Neural Networks (CNNs) and Recurrent Neural Networks (RNNs) have established to be convenient for the task, improving the performance and robustness in comparison to traditional machine learning techniques. However, some aspects must still be revisited and improved. To the best of our knowledge, this is the first article that intends to explore and propose novel gait biometric recognition systems based on Transformers, which currently obtain state-of-the-art performance in many applications. Several state-of-the-art architectures (Vanilla, Informer, Autoformer, Block-Recurrent Transformer, and THAT) are considered in the experimental framework. In addition, new configurations of the Transformers are proposed to further increase the performance. Experiments are carried out using the two popular public databases whuGAIT and OU-ISIR. The results achieved prove the high ability of the proposed Transformer, outperforming state-of-the-art CNN and RNN architectures.

\end{abstract}


\section{Introduction}
\label{sec:introduction}
\input{Introduction}

\section{Related Works}
\label{sec:relatedWork}
\input{RelatedWork}

\section{Methods}
\label{sec:proposedmethods}
\input{ProposedMethods}

%\section{Databases}
%\label{sec:databases}
%\input{Datasets}

\section{Experimental Protocol}
\label{sec:experimentalprotocol}
\input{ExperimentalProtocol}

\section{Systems Details}
\label{sec:systemConfiguration}

\input{SystemConfiguration}

\section{Experimental Results}
\label{sec:experimentalresults}

\input{ExperimentalResults}

\section{Conclusions}
\label{sec:conclusions}

\input{Conclusions}

%\section{References Section}
%You can use a bibliography generated by BibTeX as a .bbl file.
% The IEEEtran BibTeX style support page is:
 
%\section{Simple References}
%You can manually copy in the resultant .bbl file and set second argument of $\backslash${\tt{begin}} to the number of references
% (used to reserve space for the reference number labels box).

\section{Acknowledgements}
This project has received funding from the European Union’s Horizon 2020 research and innovation programme under the Marie Skłodowska-Curie grant agreement No 860315. R. Tolosana and R. Vera-Rodriguez are also supported by INTER-ACTION (PID2021-126521OB-I00 MICINN/FEDER).

%\bibliography{mybibfile}

\bibliography{mybibfile.bib}{}
\bibliographystyle{IEEEtran}

\newpage

%If you have an EPS/PDF photo (graphicx package needed), extra braces are needed around the contents of the optional argument to biography to prevent  the LaTeX parser from getting confused when it sees the complicated  $\backslash${\tt{includegraphics}} command within an optional argument. (You can create  your own custom macro containing the $\backslash${\tt{includegraphics}} command to make things simpler here.)
 
\vspace{-20pt}

\begin{IEEEbiography}[{\includegraphics[width=1in,height=1.25in,clip,keepaspectratio]{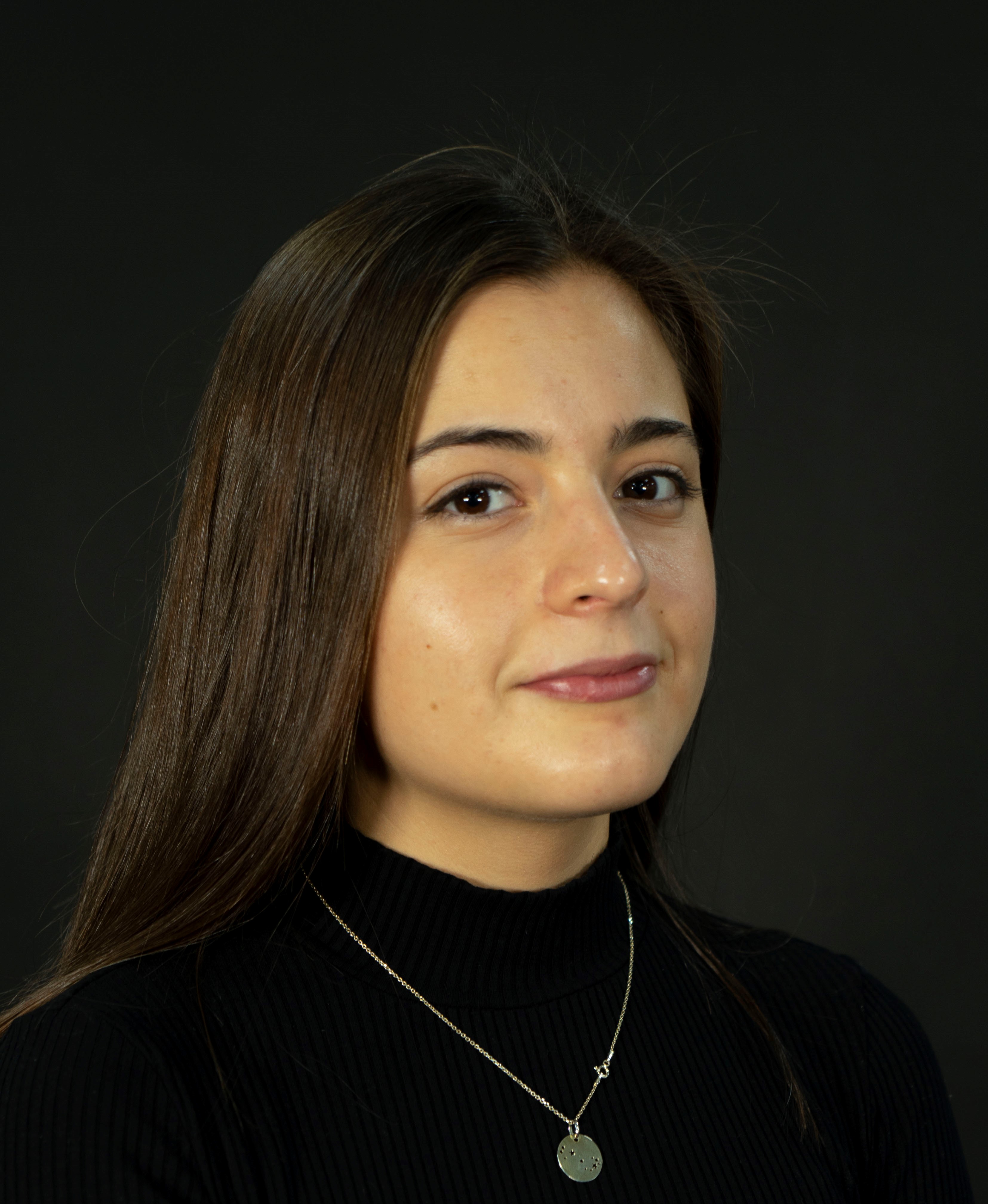}}]{Paula Delgado-Santos}
 received the M.Sc. degree in Telecommunications Engineering from Universidad Autonoma de Madrid, Spain, in 2020. At the same time, she was working in a scholarship of IBM. In 2019/2020 she was working at a Swiss University, HEIG-VD, as a Data Scientist. In 2020 she started her PhD with a Marie Curie Fellowship within the PriMa (Privacy Matters) EU project at University of Kent, U.K.. Her research interests include signal and image processing, pattern recognition, machine learning, biometrics and data protection.
\end{IEEEbiography}

\vspace{-20pt}

\begin{IEEEbiography}[{\includegraphics[width=1in,height=1.25in,clip,keepaspectratio]{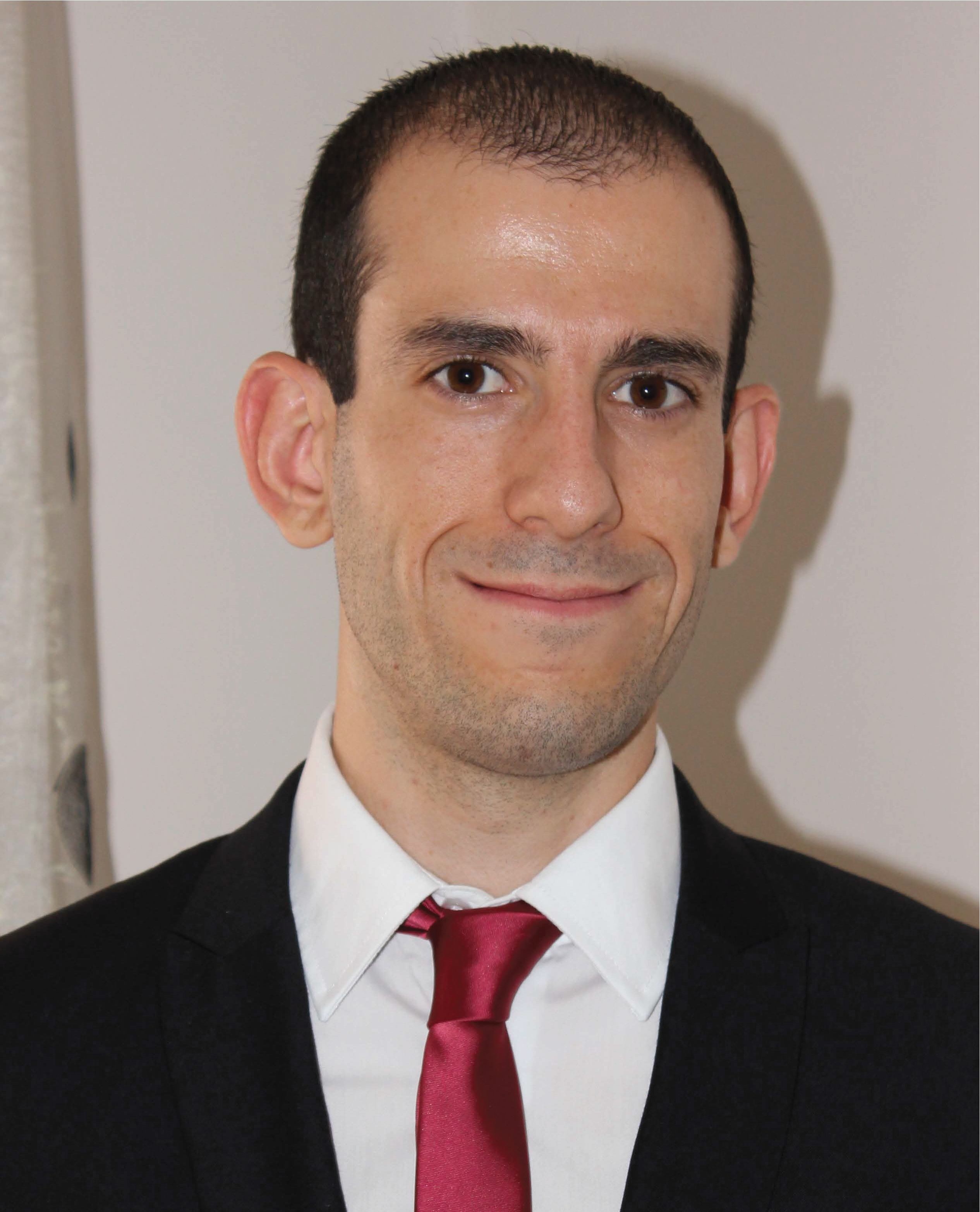}}]{Ruben Tolosana} received the M.Sc. degree in Telecommunication Engineering, and his Ph.D. degree in Computer and Telecommunication Engineering, from Universidad Autonoma de Madrid, in 2014 and 2019, respectively. In 2014, he joined the Biometrics and Data Pattern Analytics - BiDA Lab at the Universidad Autonoma de Madrid, where he
is currently collaborating as an Assistant Professor. Since then, Ruben has been granted with several awards such as the FPU research fellowship from
Spanish MECD (2015), and the European Biometrics Industry Award (2018). His research interests are mainly focused on signal and image processing, pattern recognition, and machine learning, particularly in the areas of DeepFakes, HCI, and Biometrics. He is author of several publications and also collaborates as a reviewer in high-impact conferences (WACV, ICPR, ICDAR, IJCB, etc.) and journals (IEEE TPAMI, TCYB, TIFS, TIP, ACM CSUR, etc.). Finally, he is also actively involved in several National and European projects. 
\end{IEEEbiography}

\vspace{-20pt}

\begin{IEEEbiography}[{\includegraphics[width=1in,height=1.25in,clip,keepaspectratio]{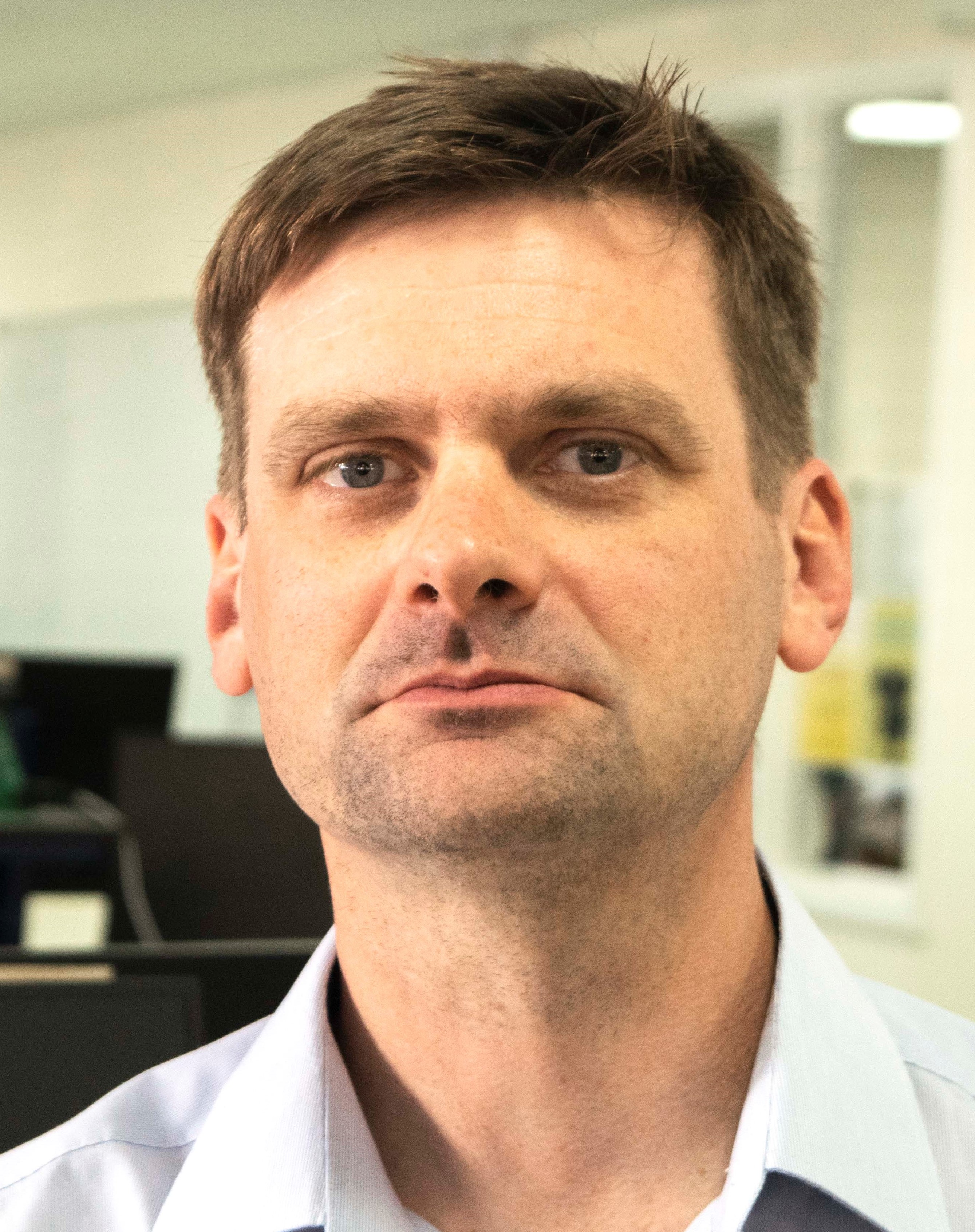}}]{Richard Guest}
 obtained his PhD in 2000. He is Professor of Biometric Systems Engineering and Head of the School of Engineering at the University of Kent. His research interests lie broadly within image processing and pattern recognition, specialising in biometric and forensic systems, particularly in the areas of image and behavioural information analysis, standardisation and mobile systems.
\end{IEEEbiography}

\vspace{-20pt}

\begin{IEEEbiography}[{\includegraphics[width=1in,height=1.25in,clip]{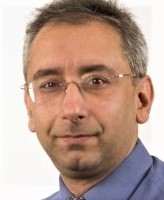}}]{Farzin Derazi}
 received the B.A. degree in Engineering Science and Economics from the University of Oxford, U.K., in 1981, the M.Sc. degree in Communications Engineering from Imperial College, U.K., in 1982, and the Ph.D. degree in Electronic Engineering from the 
University of Wales, Swansea, U.K., in 1988. He is currently with the School of Engineering and Digital Arts, University of Kent, Canterbury, U.K., where he is the Emeritus Professor of Information Engineering. His current research interests include the fields of pattern recognition and signal processing and their application in security and healthcare.
\end{IEEEbiography}

\vspace{-20pt}

\begin{IEEEbiography}[{\includegraphics[width=1in,height=1.25in,clip,keepaspectratio]{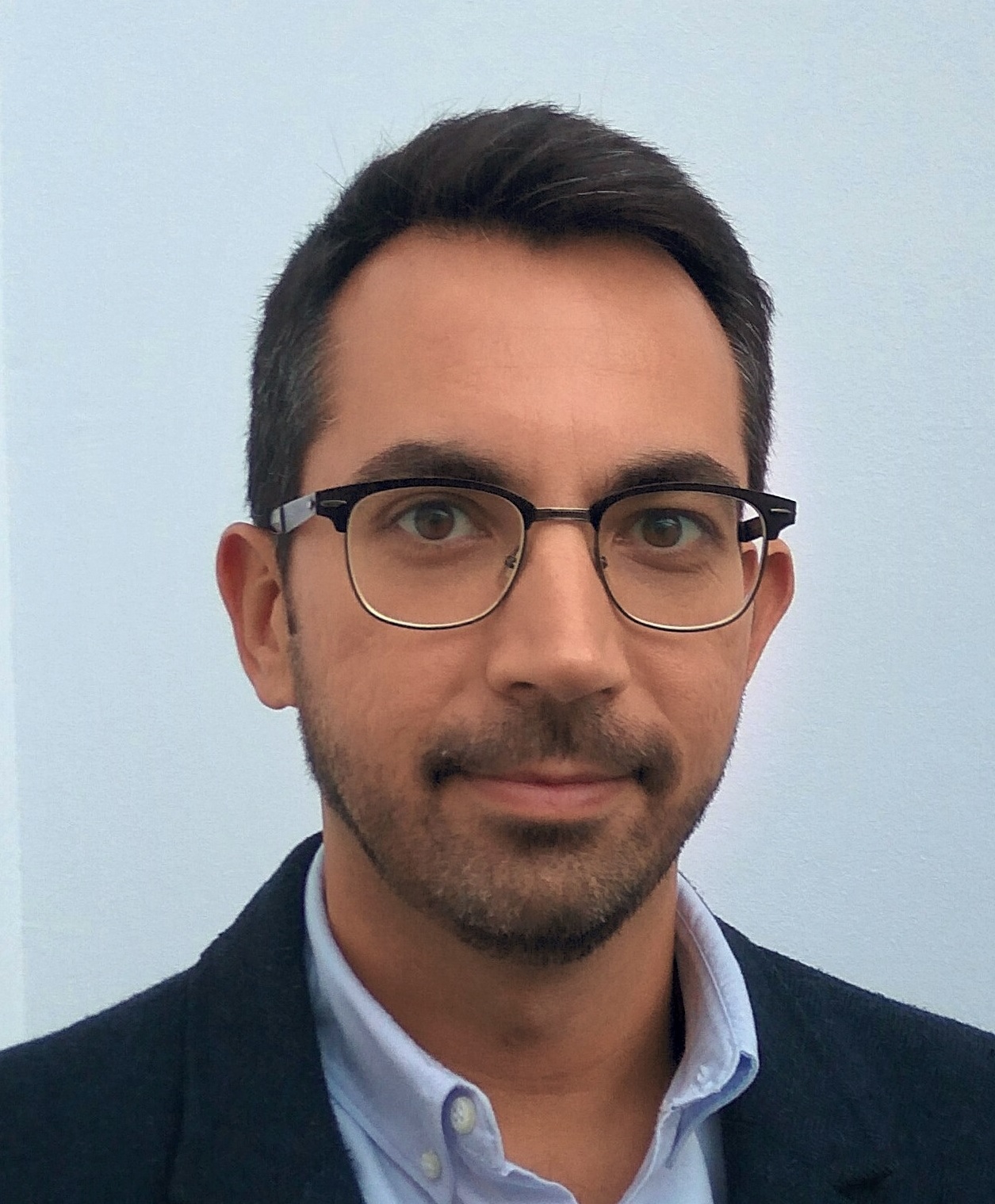}}]{Ruben Vera-Rodriguez}
 received the M.Sc. degree in telecommunications engineering from Universidad de Sevilla, Spain, in 2006, and the Ph.D. degree in electrical and electronic engineering from Swansea University, U.K., in 2010. Since 2010, he has been affiliated with the Biometric Recognition Group, Universidad Autonoma de Madrid, Spain, where he is currently an Associate Professor since 2018. His research interests include signal and image processing, pattern recognition, machine learning, and biometrics. He is the author of more than 130 scientific articles published in international journals and conferences, and 3 patents. He is actively involved in several National and European projects focused on biometrics. He has served as Program Chair for some international conferences such as: IEEE ICCST 2017, CIARP 2018 and ICBEA 2019.
\end{IEEEbiography}

%\bf{If you include a photo:}\vspace{-33pt}
%\begin{IEEEbiography}[{\includegraphics[width=1in,height=1.25in,clip,keepaspectratio]{fig1}}]{Michael Shell}
%Use $\backslash${\tt{begin\{IEEEbiography\}}} and then for the 1st argument use $\backslash${\tt{includegraphics}} to declare and link the author photo.
%Use the author name as the 3rd argument followed by the biography text.
%\end{IEEEbiography}



%\vfill

\end{document}

%% file: Introduction.tex
%BIOMETRICS has become a relevant topic for security and authentication purposes
%{B}{iometric} techniques for person recognition are now well established for security and authentication purposes
\IEEEPARstart{B}{iometrics} has become a relevant topic for security and authentication purposes\cite{jain201650}. Among the different biometric traits, gait behavioural biometrics has attracted considerable attention in recent years; for example, in surveillance scenarios where popular biometric traits such as face and fingerprint are hard or impossible to distinguish. Gait recognition uses the movement pattern of subjects by focusing on specific characteristics such as the arm swing amplitude, step frequency, and gait length~\cite{wang2003silhouette}. Depending on its specific application scenario, the gait pattern can be captured using visual sensors such as surveillance cameras~\cite{singh2018vision} or inertial sensors such as the accelerometer and gyroscope included in wearable devices~\cite{marsico2019survey}.

The popularity of gait recognition has also increased with the success of Deep Learning (DL)~\cite{sepas2022deep,filipi2022gait}. Architectures based on Convolutional Neural Networks (CNNs) and Recurrent Neural Networks (RNNs), such as Long Short-Term Memory (LSTM), have proven to be convenient for the task, improving its performance and robustness compared to traditional machine learning techniques. However, these popular DL architectures still have several disadvantages that must be revisited and improved. The main drawbacks are~\cite{vaswani2017attention,hutchins2022block}: \textit{i)} sequential computation, not allowing parallelisation within batches, \textit{ii)} compression and summarising of the previous time samples, limiting the past information seen, and \textit{iii)} vanishing gradients during back-propagation; the forget gate in a RNN removes a small portion of the previous state after each sample.
 
Transformers are more recently proposed DL architectures that have already garnered immense interest due to their effectiveness across a range of application domains such as language, vision, and reinforcement learning~\cite{tay2020efficient}. Their main advantages compared with traditional CNN and RNN architectures are~\cite{vaswani2017attention, xu2021autoformer, hutchins2022block}: \textit{i)} Transformers are feed-forward models that process all the sequences in parallel, being a more efficient technique; \textit{ii)} They apply a self-attention/auto-correlation mechanism that allows them to operate in long sequences; \textit{iii)} They can be trained efficiently even in 1 batch since all the sequence is used in every batch; and \textit{iv)} They can attend to the whole sequence, instead of summarising all the previous temporal information.

Several Transformer architectures have been recently proposed in the literature~\cite{tay2020efficient,wen2022transformers}. The original one, the Vanilla Transformer, was introduced in 2017 by Vaswani \textit{et al.}~\cite{vaswani2017attention}. It was based solely on self-attention mechanisms, dispensing with recurrence and convolutions layers entirely. Impressive results were achieved on the machine translation task, reducing also the training costs of the best models compared with the literature. Despite these improvements, the Vanilla Transformer has disadvantages for some applications based on time series: \textit{i)} the computational complexity of the attention mechanism is quadratic $O(L^2)$ where $L$ denotes the length of the input sequence; and \textit{ii)} the total memory usage is $O(N\odot L^2)$ where $N$ indicates the number of encoder/decoder layers, limiting the scalability of the model with long sequences. As a result, different Transformer architectures have recently emerged with the aim of addressing the shortcomings of the Vanilla Transformer, including: Informer~\cite{zhou2021informer}, Autoformer~\cite{xu2021autoformer}, Block-Recurrent Transformer~\cite{hutchins2022block}, and THAT~\cite{li2021two}, among others.

The present article intends to explore and propose novel behavioural biometric systems based on Transformers. The main contributions of the present study are as follows:
\begin{itemize}
    \item An in-depth analysis of state-of-the-art deep learning approaches for gait recognition on mobile devices.
    \item An overview of the main concepts of Transformers, including the key differences between popular architectures proposed in the literature. 
    \item To the best of our knowledge, this is the first study that explores the potential of Transformers for behavioural biometrics, in particular, gait biometric recognition on mobile devices. Several state-of-the-art Transformer architectures are considered in the evaluation framework (Vanilla, Informer, Autoformer, Block-Recurrent Transformer, and THAT), comparing them with traditional CNN and RNN architectures. In addition, new configurations of the Transformers are proposed to further improve the performance.
    \item An extensive experimental framework using popular public databases in gait biometric recognition. On the existing whuGAIT \cite{zou2020deep} and OU-ISIR \cite{iwama2012isir, Ngo2019OUISIR} databases, the proposed Transformer outperforms traditional CNN and RNN architectures and achieves competitive results compared with the state of the art.
\end{itemize}

The exploration and analysis included in the present study can also be very useful for other research lines, for example: \textit{i)} improving the authentication performance of other behavioural biometric traits such as handwritten signature and keystroke~\cite{2020_TIFS_BioTouchPass2_Tolosana,mondal2017person}, among many others, \textit{ii)} improving the prediction and monitoring of diseases~\cite{2021_ScientificReportsECGsMelzi}, and \textit{iii)} facilitating the training and synthesis of new data~\cite{sun2020facial,2021_AAAI_DeepWriteSYN}. 

The remainder of the article is organised as follows. Sec. \ref{sec:relatedWork} summarises previous studies in the field of gait recognition on mobile devices. Sec. \ref{sec:proposedmethods} explains the main concepts of Transformers and the key differences between the architectures considered in the study. Sec. \ref{sec:experimentalprotocol} describes the databases and experimental protocol whereas Sec. \ref{sec:systemConfiguration} provides a description of the system details. Sec. \ref{sec:experimentalresults} describes the results achieved and comparison with the state of the art. Finally, Sec. \ref{sec:conclusions} draws the final conclusions and future research lines.

%% file: RelatedWork.tex
%Works based on data fusion level.
% \usepackage{multirow}

\begin{table*}[tp!]
\centering
\caption{Summary of the most relevant methodologies for gait biometric recognition based on DL methods.}
\begin{tabular}{cccccc}
\hline
\textbf{Category}                                          & \textbf{Year} & \textbf{Ref.}                             & \textbf{Description}                   & \textbf{Performance} & \textbf{Database} \\ \hline
                                       & 2016         & \cite{gadaleta2018idnet}& CNN Feature Extractor + SVM Classifier & 92.91\%              & whuGAIT           \\ \cline{5-6}
            CNNs                                          &           &                                      &  & 44.29\%              & OU-ISIR           \\ \cline{2-6}
                                                          & 2019      & \cite{delgado2018end}                         & Fusion CNN  + Euclidean Distance                            & 92.89\%              & whuGAIT           \\ \cline{5-6}
                                                           &           &                            &                            & 40.60\%              & OU-ISIR           \\ \hline
                                      & 2020         &  \cite{watanabe2020gait}                              &      End-to-End RNN                            & 91.88\%              & whuGAIT           \\ \cline{5-6}
                                                           &           &                                   &                                       & 66.36\%              & OU-ISIR           \\ \cline{2-6}
  RNNs                                                 & 2020             & \cite{zou2020deep}                                      &                End-to-End RNN                       & 91.88\%              & whuGAIT           \\ \cline{5-6}
                                                           &           &                                       &                                       & 66.36\%              & OU-ISIR           \\ \cline{2-6}
                                                           & 2021            & \cite{tran2021multi}                                      &                   End-to-End Multi-RNN                 & 93.14\%              & whuGAIT           \\ \cline{5-6}
                                                           &           &                                      &                                      & 78.92\%              & OU-ISIR           \\ \hline
                                 & 2016  & \cite{ordonez2016deep}                            &                           Cascaded CNN + RNN              & 92.25\%              & whuGAIT           \\ \cline{5-6}
                                                         &           &                             &      & 37.33\%              & OU-ISIR           \\ \cline{2-6}
          CNNs + RNNs                                                & 2020             & \cite{zou2020deep}                                     &               2-Parallel Branches: CNN + RNN                     & 93.52\%              & whuGAIT           \\ \cline{2-6}
                                                           & 2021         & \cite{tran2021multi}                                      &                    2-Parallel Branches: CNN + Multi-RNN          & 94.15\%              & whuGAIT           \\ \cline{5-6}
                                                           &           &                                      &                               & 89.79\%              & OU-ISIR           \\ \hline
                                       &        & & \textbf{2-Parallel Branches: Temporal and Channel Modules
                      } & \textbf{94.25\%}          & \textbf{whuGAIT}          \\ 
                   \textbf{Proposed}      & \textbf{2022}   &  \textbf{Present Work} & \textbf{Temporal: Auto-Correlation + GBR CNN Layers and Recurrent Layer 
                      } &               &            \\ \cline{5-6}
                             \textbf{Transformer}                           &           &                                      & \textbf{Channel: Auto-Correlation + GBR CNN Layers} & \textbf{93.26\%}              & \textbf{OU-ISIR}          \\     
                                                                                                                  &           &                                      & \textbf{Gaussian Range Encoding in both Temporal and Channel Modules} &                &                 \\ \hline
%\multirow{CNN + Attention } & \multirow{2021}          & \multirow{\cite{huang2021lightweight} }                             &                 Lightweight Attention-Based CNN                                & 94.71\%              & whuGAIT           \\ \cline{5-6}
%                                                  &           &                                      &                                        & 97.16\%              & OU-ISIR   \\ \hline

\end{tabular}
\label{table:relatedworks}
\end{table*}
%\textcolor{red}{Use 2 subsections? 1- Gait Authentication 2- Transformers in time series}

%\subsection{Gait Biometric recognition}
Gait biometric recognition enables subjects to be authenticated based on their walking patterns. Due to the exponential increase in the number of mobile devices and the high precision of their sensors, the interest in gait recognition based on mobile devices is on the increase \cite{marsico2019survey}. One of the most popular approaches is based on the Inertial Measurement Units (IMU), e.g., accelerometer and gyroscope~\cite{sprager2015inertial}. Table \ref{table:relatedworks} provides a summary of the most relevant methodologies for gait biometric recognition on mobile devices based on DL methods. It is important to highlight that all approaches consider the same experimental protocol proposed in~\cite{zou2020deep} for two popular public databases in the literature: \textit{i)} whuGAIT~\cite{zou2020deep}, which comprises accelerometer and gyroscope data acquired from mobile devices, and \textit{ii)} OU-ISIR~\cite{iwama2012isir,Ngo2019OUISIR}, which includes accelerometer and gyroscope data obtained from IMU sensors.

%evaluated with the experimental protocol proposed in \cite{zou2020deep}.

In the past few years, the research community has focused on DL models to improve the robustness of gait recognition systems, extracting more discriminative features. As both the spatial and temporal information of the gait pattern is important for the task, DL architectures based on CNN and RNN have been utilised. One of the earliest systems based on DL models using CNNs was created by Gadaleta and Rossi in \cite{gadaleta2018idnet}. The authors used CNNs for feature extraction and a Support Vector Machine (SVM) for the final classification with 0.15\% misclassification rates. The score was obtained in less than five walking cycles with their own collected database. Their results proved how DL methods could extract more discriminative features compared with previous machine learning methods. The same model was evaluated in \cite{zou2020deep} following a predefined experimental protocol, obtaining an accuracy of 92.91\% in the whuGAIT database\cite{zou2020deep}, and 44.29\% accuracy in the OU-ISIR database~\cite{iwama2012isir, Ngo2019OUISIR}. Another approach based on CNNs was presented by Delgado-Esca\~{n}o \textit{et al.} in \cite{delgado2018end}, dividing the data into two branches, according to each sensor (accelerometer and gyroscope). The output of both branches were concatenated to produce a joint feature vector. Cross-validation was used, achieving 95.20\% accuracy with the OU-ISIR database using an internal experimental protocol. Following the predefined experimental protocol presented in \cite{zou2020deep}, results of 92.89\% and 44.29\% accuracy were achieved in the whuGAIT and OU-ISIR databases, respectively. However, by using only CNNs, the system focuses mainly on spatial characteristics, leaving out the temporal information. 

To overcome this drawback, RNNs were proposed, extracting temporal features from the time sequences. Watanabe \textit{et al.} created an end-to-end RNN with a softmax layer~\cite{watanabe2020gait}. The model was tested with the experimental protocol presented in \cite{zou2020deep}, achieving a 91.88\% accuracy with whuGAIT database, and 66.36\% accuracy with OU-ISIR database. Zou \textit{et al.} evaluated RNNs in \cite{zou2020deep} over the OU-ISIR database achieving 78.92\% accuracy. They also presented the whuGAIT database and proposed a predefined experimental protocol, achieving 93.14\% accuracy.

Hybrid approaches have also been proposed in the literature, trying to achieve a more complex structure, where the CNN extracts spatial features while the RNN obtains temporal features. Ordo\~{n}ez and Roggen presented in \cite{ordonez2016deep} DeepConvLSTM, which comprises convolutional layers, followed by recurrent and softmax layers. The model obtained 95.8\% F1-score for the activity recognition task with the Opportunity database. The system was also evaluated for gait recognition in \cite{zou2020deep}, achieving 92.25\% and 37.33\% accuracy for the whuGAIT and OU-ISIR databases, respectively. Also, Zou \textit{et al.} presented in \cite{zou2020deep} an hybrid approach with two-parallel branches, one CNN and one RNN. The extracted features were independent in each branch, obtaining a view of the raw data with both convolutional and recurrent layers. After each branch, the features were concatenated and fed into a fully connected layer. The authors achieved 93.52\% accuracy on the presented whuGAIT database.

% also in [DLBGRUSW] a paper that with a comparison of CNNs and RNNs for HAR
%Include GANs (in [DLBGRUSW])
%Yu \textit{et al.} created a model based on Generative Adversarial Networks (GANs). The aim of the authors was minimizate the influence of view angle, weight and cloting
Previous approaches are based on gait cycle detection first. The input of the DL models is an interval time between two consecutive occurrences of the gait pattern, i.e., putting the same foot on the ground \cite{marsico2019survey}. Gait cycle detection is usually a tedious task that can induce to errors due to the sensor restrictions (e.g., noise-sensitive, sensor specification, body placement, etc.). To solve this problem, Tran \textit{et al.} proposed in \cite{tran2021multi} a new approach using window-based data segment. The authors used a Multi-RNN model considering fixed-length segments as input, without the need to extract gait cycles. The authors achieved an accuracy of 93.14\% for the whuGAIT database, and 78.92\% for the OU-ISIR database. In addition, the same authors introduced an hybrid approach, achieving 94.15\% and 89.79\% accuracy for the whuGAIT and OU-ISIR databases, respectively. 

%Attention mechanism has emerged to solve the problem of long-term dependency of RNNs. Huang \textit{et al.} in \cite{huang2021lightweight} presented an attention-based CNN model based on contextual encoding information. The model achieved 94.71\% of accuracy in the whuGAIT database and 97.16\% of accuracy in the OU-ISIR database.

Despite the success of CNN and RNN architectures, some of their limitations could still be revisited and improved such as the limited window size for RNNs. By summarising all previously observed information into one vector, these approaches miss temporal information that is relevant to gait biometric recognition. Due to the limitations highlighted, this article explores the potential of recently developed Transformer architectures for gait biometric recognition and proposes new configurations to further improve the results. Table~\ref{table:relatedworks} includes the results achieved using our proposed Transformer.

%% file: ProposedMethods.tex
%\begin{figure}[tp]
%\begin{center}
%   \includegraphics[trim={0cm 0cm 0 0}, width=.7\linewidth]{VanillaTransformer.pdf}
%\end{center}
%   \caption{Graphical representation of the Encoder of the Vanilla Transformer~\cite{vaswani2017attention}.}
%\label{fig:VanillaTransformer}
%\end{figure}

%Transformers
%desde el primero al ultimo que utilice
%explicarlo bien con figuras

This section provides an overview of the main concepts of Transformers, including the key differences between recent architectures proposed in the literature. To facilitate the understanding of this section, we include in Fig. \ref{fig:TransformerArchs} a graphical representation of the different Transformer architectures. As the present article is related to behavioural recognition, we focus only on the encoder part of the Transformer.

%The original Transformer~\cite{vaswani2017attention}, Vanilla Transformer,  is reviewed together with several structure variations focusing on time sequences (Informer \cite{zhou2021informer}, Autoformer \cite{xu2021autoformer}, Block-Recurrent Transformer\cite{hutchins2022block}, and THAT \cite{li2021two}). For this study exclusively the encoder part is considered, as shown in Fig. \ref{fig:VanillaTransformer}.

%Classification Transformers typically use a simple encoder structure, where the Self-Attention sublayer performs representation learning, and the feed-forward layer generates probabilities for each class.

\subsection{Vanilla Transformer}

\begin{figure*}[tp]
    \begin{center}
       \includegraphics[trim={0cm 0cm 0 0}, width=0.95\linewidth]{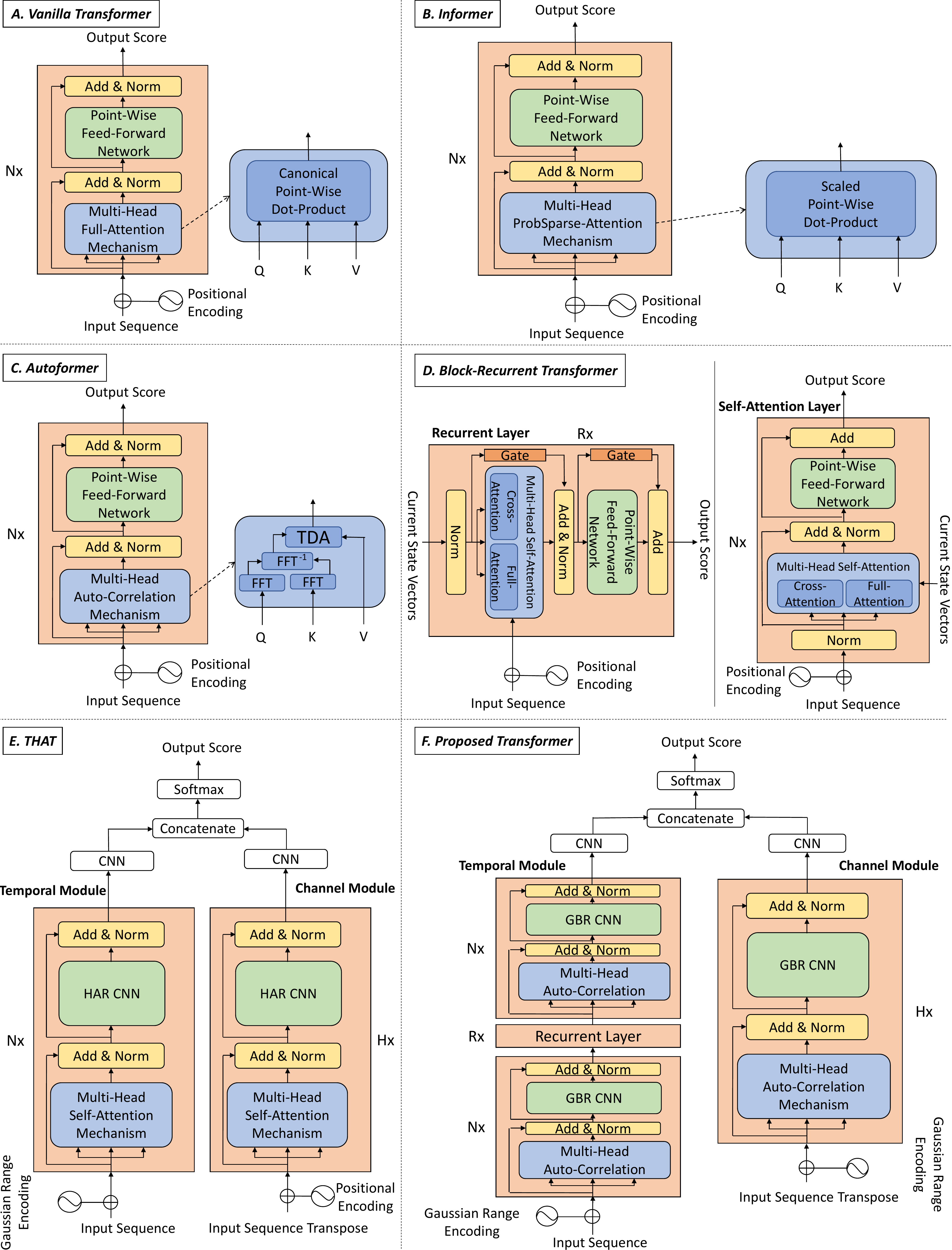}
    \end{center}
    \caption{Graphical representation of the Transformer architectures used in this study (Vanilla Transformer~\cite{vaswani2017attention}, Informer~\cite{zhou2021informer}, Autoformer~\cite{xu2021autoformer}, Block-Recurrent~\cite{hutchins2022block}, THAT~\cite{li2021two}, and the proposed one). Q: Queries; K: Keys; V: Values; Nx,Hx,Rx: they refer to the number of layers of each type; FFT: Fast Fourier Transform; TDA: Time Delay Aggregation; HAR CNN: Human Activity Recognition CNN; GBR CNN: Gait Biometric Recognition CNN.}
    \label{fig:TransformerArchs}
\end{figure*}

The original Vanilla Transformer was presented in~\cite{vaswani2017attention} for the task of machine translation. It was defined as a multi-layer encoder-decoder architecture with no recurrence and convolution layers. Fig. \ref{fig:TransformerArchs} \textit{A.} provides a graphical representation of the encoder, which is composed of a stack of $N$ identical layers. Each layer is mainly formed by two different sub-layers: \textit{i)} a multi-head Self-Attention mechanism (Full-Attention), and \textit{ii)} a point-wise feed-forward network. Subsequent of each sub-layer, a residual connection and a layer normalisation are considered ($Add~\&~Norm$ in Fig. \ref{fig:TransformerArchs}). The input sequence is a matrix $X ~\epsilon~ \mathbb{R}^{c \times L}$ where $c$ is the number of channels and $L$ the length of the sequence.

The encoder maps each position $l$ of the input sequence $X=(x_0,x_1,...,x_l,...,x_L)$ into hidden states $Z=(z_0,z_1,...,z_l,...,z_L)$. The output of each sub-layer is $LayerNorm (X + Sublayer(X))$, where $Sublayer(X)$ is the function implemented by the multi-head Self-Attention mechanism or the point-wise feed-forward network. Both the input $X$ and output $Z$ have the same dimension $L$ to facilitate the work of the residual connections. As no recurrence and convolutional layers are considered in the Vanilla Transformer, a previous encoding of the model is needed to keep certain information about the position $l$ of the sample in the input sequence. This is achieved using a positional encoding placed at the input of the model.  

We describe next the key aspects of the positional encoding, multi-head Self-Attention mechanism, and the point-wise feed-forward network for a better understanding of the Vanilla Transformer, and the later Transformer implementations.

\begin{figure*}[htp]
\begin{center}
   \includegraphics[trim={0cm 0cm 0 0}, width=1\linewidth]{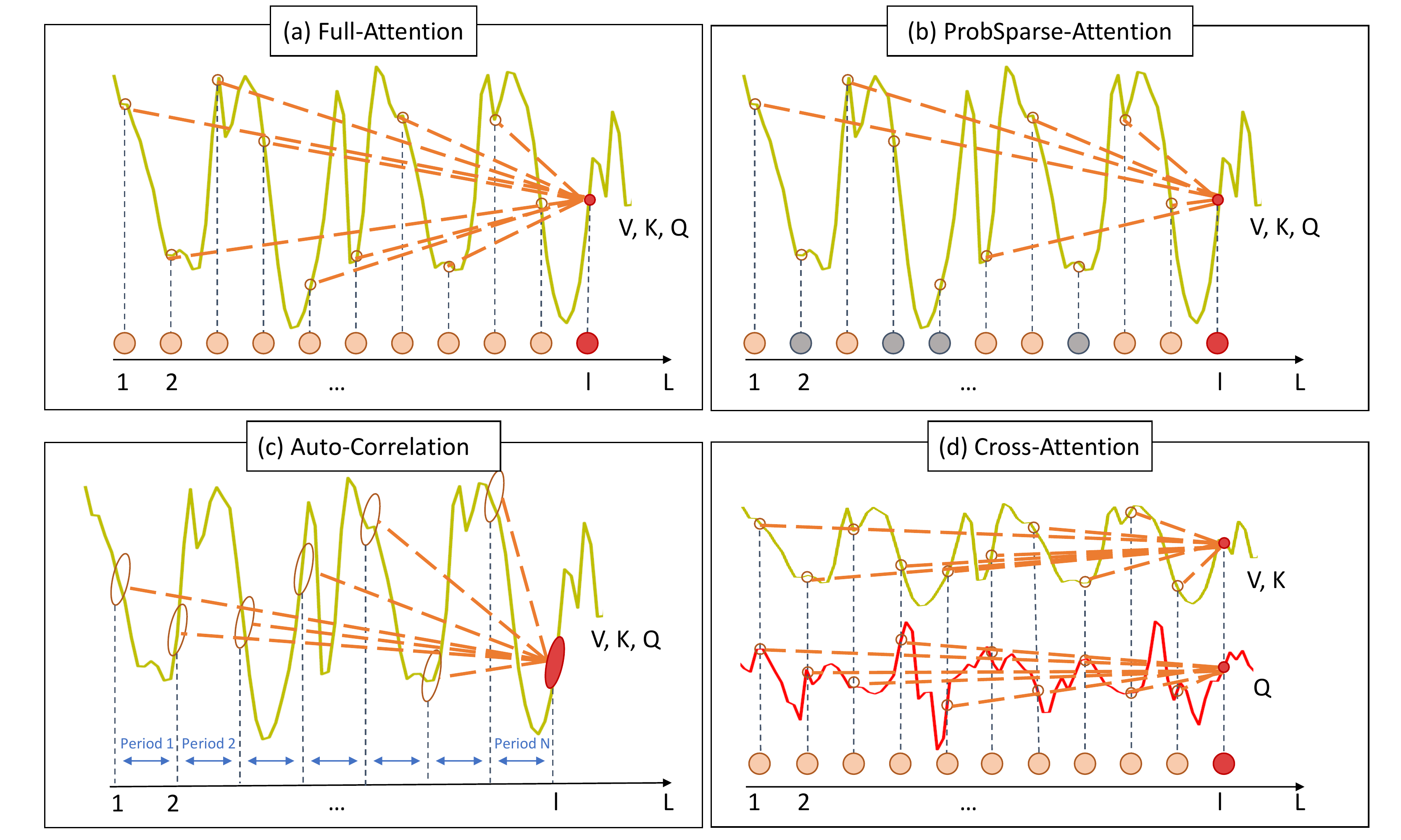}
\end{center}
   \caption{Graphical representation of Attention and Auto-Correlation mechanisms. (a) Full-Attention (Vanilla Transformer~\cite{vaswani2017attention}); (b) ProbSparse-Attention (Informer \cite{zhou2021informer}); (c) Auto-Correlation (Autoformer \cite{xu2021autoformer}); and (d) Cross-Attention (Block-Recurrent Transformer \cite{hutchins2022block}). The solid line represents the input sequence and the red one (second line) the recurrent states in Cross-Attention. The red points/series are the position $l$ of the sequence of length $L$ with $V$ values, $K$ keys, and $Q$ queries. The orange points represent the mapped points/series along the entire sequence, while the grey ones are points not mapped. Figure adapted from~\cite{xu2021autoformer}.}
\label{fig:AttentionModules}
\end{figure*}

\subsubsection{Positional Encoding}

It encodes the relative and/or absolute position of the sample $l$ of the input sequence. In the original work, Vaswani \textit{et al.} \cite{vaswani2017attention} preserved the relative context using a fixed point encoding with the sine and cosine functions:
\begin{equation}
\begin{split}
    PE_{(pos, 2l)} = sin(pos/10000^{2l/L}) \\
    PE_{(pos, 2l+1)} = cos(pos/10000^{2l/L})
\end{split}    
\end{equation}

where \textit{pos} is the position and $l$ the length. The positional encoding has the same length $L$ as the embeddings, so that the two can be summed. The output of the positional encoding is:
\begin{equation}
    \hat{x_{l}} = x_{l} + PE_{(l)}
\end{equation}

\subsubsection{Multi-Head Self-Attention Mechanism}
This mechanism is responsible for mapping scattered points along the entire sequence, studying the long-range dependencies. This mechanism avoids the limited time window problem of the previous approaches (e.g., RNNs). 
The information aggregation is accomplished with a Full-Attention mechanism where the outputs are the weighted sum of the values $V$ according to the canonical point-wise dot-product of the queries $Q$ with the corresponding keys $K$. Fig. \ref{fig:AttentionModules} (a) provides a graphical representation of the Full-Attention mechanism. The solid line represents the input sequence with its values $V$, keys $K$, and queries $Q$. The red point represents the position $l$ in the sequence with length $L$. The orange points are the scattered points mapped in the Full-Attention mechanism for the red point at position $l$. The Full-Attention mechanism can be defined as:
\begin{equation}
    Attention(Q,K,V)=softmax(\frac{QK^{T}}{\sqrt{d_{k}}})V
\end{equation}
where $d_k$ is the dimension of the queries $Q$ and keys $K$, and $\sqrt{d_k}$ is a scaling factor that enables flatter gradients. $Q=XW_Q$, $K=XW_K$, $V=XW_V$ are the linear projections of $X$ in the corresponding projection parameters $d_k$, $d_k$, and $d_v$ respectively where $W_Q ~\epsilon~ \mathbb{R}^{{L\times d_k}}$, $W_K ~\epsilon~ \mathbb{R}^{{L\times d_k}}$, and $W_V ~\epsilon~ \mathbb{R}^{{L\times d_v}}$. The computational cost is quadratic $O(L^2)$ where $L$ denotes the length of the input sequence.

Alternatively to apply one single projection of the queries, keys, and values, better results can be achieved with $h$ independent projections to $d_k$, $d_k$, and $d_v$ respectively. The multi-head Self-Attention is based on a concatenation and final projection of the $h$ independent heads:
\begin{equation}
    MultiHead(Q,K,V)=[head_1, ..., head_h]W^O
\end{equation}
where $head_i = Attention(Q_i, K_i, V_i)$ and $W^O ~\epsilon~ \mathbb{R}^{{hd_v\times L}}$ is the final matrix. To achieve the same length $L$ of the input sequence, $d_v = L/h$.

\subsubsection{Point-Wise Feed-Forward Network}
In addition to the multi-head Self-Attention sub-layer, the Vanilla Transformer has a point-wise feed-forward network. This consists of two linear transformations with a ReLU activation in between, operating in each position independently. The input and output dimensions are the same, $L$.

%The decoder part of a Vanilla Transformer would have a similar architecture, but using Cross-Attention in the multi-head self-attenuation mechanism. The Cross-Attention has the same architecture than the Full-Attention mechanism described before. Fig \ref{fig:AttentionModules} (c) shows how Cross-Attention uses two input streams, one is the input sequence, and the other the embeddings after roasting by the encoder. In this way, Cross-Attention uses the values and keys of the first sequence, input sequence, and the queries of the second sequence, embeddings.

To summarise, the Vanilla Transformer has shown great advances in Natural Language Processing and Computer Vision applications but still needs to be adapted for time sequences. Aspects such as the periodicity or seasonality, and long- and short-range dependencies still need to be revisited \cite{wen2022transformers}. To alleviate these drawbacks, different Transformers have been proposed in the research community, modifying aspects such as the multi-head Self-Attention sub-layer and the positional encoding. 

%From the time sequences forecasting field, the structure of the Vanilla Transformer has been adapted and improved. As the authors were working with a prediction task, the complete encoder-decoder structure has been modified, as well as the positional encoding. Since recognition is a classifier-based task, only the encoder modifications will be taken into account. In addition, not having calendar timestamp data as in the prediction task, the adaptation of the proposed timestamp encodings becomes impossible.

\subsection{Informer}

Zhou \textit{et al.} presented in \cite{zhou2021informer} a new Transformer architecture named Informer. Informer is an adaptation of the Vanilla Transformer for Long Sequence Time-series Forecasting (LSTF). Some limitations of the Vanilla Transformer are the quadratic time complexity $O(L^2)$, the high memory usage $O(N \odot L^2)$ with $N$ encoder layers, and the inherent limitation of the encoder-decoder architecture. To overcome these drawbacks, the authors proposed several improvements. The multi-head Self-Attention mechanism based on Full-Attention was changed by ProbSparse-Attention to scattered points, as provides Fig. \ref{fig:TransformerArchs} \textit{B}. The Full-Attention to the input sequence is reduced to half, more favourable handling long-range sequences. The canonical dot-product was replaced by a scaled dot-product. Informer reduces the time complexity to $O(L~log~L)$ and the memory usage to $O(L~log~L)$. In addition, previous studies have shown  a potential sparsity in Full-Attention. As a result, the authors decided to use a selective strategy on all probabilities, i.e., Sparse-Attention \cite{child2019generating} (sparsity coming from separate spatial correlations) and LogSparse-Attention \cite{li2019enhancing} (selecting points through exponential increasing intervals). Fig. \ref{fig:AttentionModules} (b) provides a graphical representation of the ProbSparse-Attention mechanism. The solid line denotes the input sequence with the extracted values $V$, keys $K$, and queries $Q$. The red point represents the position $l$ in the input sequence. The ProbSparse-Attention mechanism, unlike the Full-Attention mechanism that looks at all previous points, chooses selected dominant points (orange) in the input sequence, while the grey ones are not used.
%In Full-Attention the probability $p(k_j|q_i)$ is similar to a uniform distribution, resulting the Self-Attention layer a redundant sum of values $V$. Zhou \textit{et al.} in \cite{zhou2021informer} introduced ProbSparse-Attention allowing every key to only attend $u$ dominant queries as can be seen in Fig \ref{fig:AttentionModules} (b). The authors use the similarity between distribution $p$ and $q$ to determinate the dominant queries. Kullback-Leiber divergence was used where:
%\begin{equation}
%    KL(q||p) = ln\sum_{l=1}^{d_k}e^{\frac{q_ik_l^T}{\sqrt{d_k}}} - \frac{1}{LK}\sum_{l=1}^{d_k}{\frac{q_ik_j^T}{\sqrt{d_k}}}-ln(d_k)
%\end{equation}
%Leaving aside the constant $ln(d_k)$, the sparsity measurement of the query %$i$ is defined as:
%\begin{equation}
%    M(q_i,K) = ln\sum_{l=1}^{d_k}e^{\frac{q_ik_l^T}{\sqrt{d_k}}} - \frac{1}{LK}\sum_{l=1}^{d_k}{\frac{q_ik_j^T}{\sqrt{d_k}}}
%\end{equation}
%where the first term is in the Log-Sum-Exp (LSE) of $q_o$ with all keys and the second term is in its arithmetic mean. The larger the $M(q_i,K)$ term, the more diverse the probability of Attention $p(k_j|q_i)$ will be, and therefore the more likely it will tend to contain dominant point-wise dot-product pairs.

\subsection{Autoformer}
Autoformer was presented by Wu \textit{et al.} in \cite{xu2021autoformer} for the task of long-term forecasting. In this Transformer architecture, the original multi-head Self-Attention mechanism based on Full-Attention was changed by Auto-Correlation.  %The Auto-Correlation block have some improvements compared with a normal Self-Attention block (i.e., the measurement of the time-delay similarity between the inputs, and aggregated top-k similar sub-series to produce the output). %with a reduced complexity of O(LlogL).
Contrary to previous Transformers, where the proposed dot-product only establishes point connections, the Auto-Correlation mechanism not only goes over long-range dependencies but also periodicity-based dependencies. Using series-wise connections instead of point-wise, Autoformer achieves a time complexity of $O(L~log~L)$, and breaks the information utilisation bottleneck. Fig. \ref{fig:AttentionModules} (c) shows a graphical representation of Auto-Correlation. It takes into consideration series of points in the same position during previous periods of the input sequence instead of scattered points.

Fig. \ref{fig:TransformerArchs} \textit{C.} provides a graphical representation of Autoformer. The multi-head Auto-Correlation sub-layer comprises two main sub-blocks: \textit{i)} an aggregated top-k similar sub-series, calculated by Fast Fourier Transform (FFT) and based on periodicity (instead of scattered points like the Self-Attention family), and \textit{ii)} Time Delay Aggregation (TDA) among periods (instead of point-wise dot-product like in the Self-Attention family), used for the information aggregation.

The \textit{aggregated top-k similar sub-series} presents series-wise connections based on period-based dependencies. The sub-series are correlated between them at the same position in previous periods, which are congenitally sparse. For an input sequence $X=(x_0,x_1,...,x_l,...,x_L)$,  $X ~\epsilon~ \mathbb{R}^{c \times L}$ where $c$ is the number of channels and $L$ the length of the input sequence, the Auto-Correlation $R_{XX}(\tau)$ can be obtained by FFT based on Wiener-Khinchin theorem as:
\begin{align}
    S_{XX}(f)& = FFT(X)FFT^*(X)
    \\
    R_{XX}(\tau)& = FFT^-1(S_{XX}(f))
\end{align}
where $FFT^*$ is the conjugate operation, $FFT^{-1}$ its inverse, and $S_{XX}(f)$ is the Auto-Correlation obtained in the frequency domain.
%For an input sequence ($X=(x_0,x_1,...,x_l,...,x_L)$), $l$ is the length of the sequence, the Auto-Correlation $R_X_X(\tau)$ can be obtained as:
%\begin{equation}
%    R_{XX}(\tau)=\lim_{L\to\infty}\frac{1}{L}\su%m_{l=1}^{L}X_nX_{l-\tau}
%\end{equation}
%where $R_{XX}(\tau)$ is the time-delay between $X_n$ and its $\tau$ delay series $X_{l-\tau}$ and $\tau$ the estimated period length. Being $k$ the most possible period lengths factor, the selected time delays are $\tau_1,...,\tau_k$. The period-based dependencies are predicted by weighting the corresponding Auto-Correlation.

The \textit{Time Delay Aggregation (TDA)} sub-block links the sub-series over the selected time delays $\tau_1,...,\tau_k$. This operation aligns sub-series in the same phase of the predicted periods, contrary to point-wise dot-product in the Self-Attention family. Finally, the sub-series are aggregated by softmax normalised function. The Auto-Correlation mechanism can be defined as:
\begin{equation}
\begin{split}
    \tau_1, ...,\tau_k = \underset{\tau~\epsilon~ (1,...,L)}{argTopK} (R_{Q,K}(\tau)) \nonumber  
\end{split}
\end{equation}
\begin{equation}
\begin{split}
    \hat{R}_{Q,K}(\tau_1),...,\hat{R}_{Q,K}(\tau_k) =SoftMax(R_{Q,K}(\tau_1),..., R_{Q,K}(\tau_k)) \nonumber
\end{split}
\end{equation}
\begin{equation}
\begin{split}
    Auto-Correlation(Q,K,V) =\sum_{i=1}^{k}Roll(V, \tau_i)\hat{R}_{Q,K}(\tau_i)
\end{split}
\end{equation}

where ${argTopK}$ takes the output of $topK$ Auto-Correlations along $l$, $R_{Q,K}$ is the Auto-Correlation between $Q$ and $K$ series, and $Roll(V, \tau_i)$ scroll $X$ with a $\tau$ time delay, re-introducing the elements moved beyond the first position to the last one. 

\subsection{Block-Recurrent Transformer}

Hutchins \textit{et al.} introduced the Block-Recurrent Transformer in \cite{hutchins2022block} for the task of auto-regressive language modelling. This Transformer introduces a recurrent form of attention. It is presented as an alternative to using the dot-product or periodicity-based series mechanism, which fix an attention window size. The Block-Recurrent Transformer summarises the sequence that the model has previously seen. The time complexity is linear $O(L)$. The recurrent layers operate on series-wise connections as in the Autoformer, achieving linear memory consumption $O(L)$. The Block-Recurrent Transformer is based on a sliding-window attention mechanism \cite{beltagy2020longformer}. Given an input $X$ with length $L$, a causal mask is applied by a sliding window with size $W$ where every sample can attend only to the previous $W$ samples. Being the attention matrix of Full-Attention $L \times L$, the Block-Recurrent Attention matrix is $W \times W$, where $W << L$. The sliding-window attention processes multiple blocks of size $W$ at the same time. 

Fig. \ref{fig:TransformerArchs} \textit{D.} provides a graphical representation of the Block-Recurrent Transformer architecture, which comprises two main directions: \textit{i)} vertical direction (Self-Attention Layer in Fig.\ref{fig:TransformerArchs} \textit{D.}), where layers are placed in the usual way; and \textit{ii)} horizontal direction (Recurrent Layer in Fig.\ref{fig:TransformerArchs} \textit{D.}), where layers contain recurrence. Both directions attend to the input sequence $X$ and to the recurrent states $S$.

The \textit{vertical direction} presents a multi-head Self-Attention sub-layer with two attentions: \textit{i)} Full-Attention to the input sequence $X$, Fig. \ref{fig:AttentionModules}(a), and \textit{ii)} Cross-Attention to the recurrent states $S$, which are initialised to 0, to extract the queries $Q$ whereas the keys $K$ and values $V$ are extracted from the input sequence $X$, Fig. \ref{fig:AttentionModules}(d). 

The \textit{horizontal direction} also presents a multi-head Self-Attention sub-layer with two attentions: \textit{i)} Cross-Attention to the input sequence $X$ to extract the queries $Q$ while the keys $K$ and values $V$ are extracted from the recurrent states $S$, Fig. \ref{fig:AttentionModules}(d), and \textit{ii)} Full-Attention to the recurrent states $S$, Fig. \ref{fig:AttentionModules}(a). The horizontal direction applies recurrence where the residual connections are replaced by gates, allowing the model to forget. Also, the gates help the model to apply Full-Attention and Cross-Attention in parallel. For the recurrence, the current states $S$ is going to be modified by residual connection gates. The input of the state at the next window ($s_{w+1}$) depends on the output of the state at the actual window ($s_{w}$):
\begin{equation}
\begin{split}
    s_{w+1} = s_w \odot g + z_w \odot  (1-g) \nonumber
\end{split}
\end{equation}
\begin{equation}
\begin{split}
   g = \sigma (b^{(g)})
\end{split}
\end{equation}
\begin{equation}
\begin{split}
    z_w = W^{(z)}h_w + b^{(z)} \nonumber
\end{split}
\end{equation}

where $\odot$ is the point-wise multiplication, $g$ the gate, $z_w$ the learned convex combination, $b^{(g)}$ and $b^{(z)}$ are trainable bias vectors (learned functions between the distance of the query $Q$ and key $K$), $W$ the weight matrix, $h_w$ the output of the corresponding sub-layer (i.e., multi-head Self-Attention mechanism or point-wise feed-forward network), and  $\sigma$ the sigmoid function.

The Block-Recurrent Transformer applies layer normalisation before the multi-head Self-Attention sub-layer, and before the point-wise feed-forward network. Dropout is also introduced before the multi-head Self-Attention sub-layer and after the point-wise feed-forward network. 

\subsection{THAT}

Contrary to images, which have spatial information in two dimensions (2D), temporal sequences might consider spatial information in one dimension (1D) in each time position. Furthermore, they can extract temporal information for each time position in a second dimension. The spatial information is available in the same way, between the different channels of each time sample, which can be called as channel-over-time features. On the other hand, being a temporal sequence, there are time-over-channel features, which need to be treated as a temporal sequence. 

Based on this idea, the \textit{Two-stream Convolution Augmented Human Activity Transformer} (THAT) model was proposed by Li \textit{et al.} in \cite{li2021two}. The authors proposed a new Transformer architecture for Human Activity Recognition (HAR). Fig. \ref{fig:TransformerArchs} \textit{E.} provides a graphical representation of the THAT Transformer. The model contains two parallel modules for the feature extraction: \textit{i)} Temporal Module (in charge of time-over-channel features), and \textit{ii)} Channel Module (in charge of channel-over-time features). Subsequently, all extracted features are concatenated for the prediction task. 

%In the Channel Module, since the data are represented as channel-over-time, spatial patterns are extracted. In this Module, the Vanilla Transformer architecture is used. On the contrary, the Temporal Module need adaptation considering time-over-channel features. 

The authors claimed that the original positional encoding considered in the Vanilla Transformer~\cite{vaswani2017attention} might not be sufficient to capture all the temporal information along the sample as it is defined on a single point. As a result, the authors proposed a Gaussian range encoding, suggesting the use of a range of points rather than just one. Furthermore, several ranges $g$ can be used at the same time, allowing to have different contexts of the sample $x_{l}$. 

Assuming $g~\epsilon~\mathbb{R}^G$ different ranges, $\mathcal{N}(\mu^g,\,\sigma^{g}) ~\epsilon~ \mathbb{R}^{L\times G} $ is a Gaussian distribution with the probability $p^g(l)$. Being $p_l=( \frac{p^1(l)}{\zeta},...,\frac{p^G(l)}{\zeta})$ the distribution over the $G$ ranges with a normalisation factor $\zeta$, $V=(v_1,...,v_G)$ is the values vector over the ranges. All $\mu$, $\sigma$, and $V$ variables are initialised randomly and re-adjusted with the training of the whole model. To summarise, the output of the Gaussian range encoding at position $l$ is:

\begin{equation}
    \hat{x_{l}}= x_{l} + V^Tp_l
\end{equation}

In addition, as the point-wise feed-forward layer proposed in the Vanilla Transformer~\cite{vaswani2017attention} focuses attention on a single point in time, the authors implemented a multi-scale CNN with adaptive Scale-Attention in both Temporal and Channel Modules. They replaced the linear transformations of the original feed-forward layer with a HAR CNN. Also, by introducing Scale-Attention Adaptive, the training can be adjusted to the different ranges introduced by the Gaussian range encoding.   

%\begin{figure}[tp]
%\begin{center}
%   \includegraphics[trim={0cm 0cm 0 0}, width=1.\linewidth]{ProposedTransformer_noframe.pdf}
%\end{center}
%   \caption{Proposed Transformed-based architecture.}
%\label{fig:ProposedTransformer}
%\end{figure}

\subsection{Proposed Transformer}

Finally, Fig \ref{fig:TransformerArchs} \textit{F.} presents the new proposed Transformer based on a selection of the best components presented in previous Transformer architectures. First, we consider a parallel two-stream architecture with Temporal and Channel Modules, similar to the THAT approach presented in \cite{li2021two}. Unlike the THAT approach, we consider a Gaussian range encoding as input of both Temporal and Channel Modules. In addition, for the Temporal Module (left branch), we consider a combination of multi-head Auto-Correlation layers, proposed in Autoformer~\cite{xu2021autoformer}, and a recurrent layer in between, proposed in Block-Recurrent Transformer~\cite{hutchins2022block}. For the multi-head Auto-Correlation layer, we design a specific multi-scale Gait Biometric Recognition (GBR) CNN sub-layer. Regarding the Channel Module (right branch), we consider a multi-head Auto-Correlation sub-layer together with a multi-scale GBR CNN sub-layer. After each sub-layer, a residual connection is applied followed by a normalisation of the layer, similar to the Vanilla Transformer~\cite{vaswani2017attention}.

%Future: residual connections: gates, work 95....\%

%% file: ExperimentalProtocol.tex
%\textcolor{red}{NUMBER OF SAMPLES OF EACH USER FOR EACH DATASET: TRAINING, VALIDATION, EVALUATION}

Two popular public databases used for research in gait recognition on mobile devices are considered in the evaluation framework of the present study: \textit{i)} whuGAIT~\cite{zou2020deep}, and \textit{ii)} OU-ISIR~\cite{Ngo2019OUISIR}. These databases have been selected as they also contain predefined experimental protocols (i.e., development and evaluation datasets), allowing for a fair comparison between state-of-the-art approaches.

\subsection{WhuGAIT database}
The whuGAIT database was introduced in \cite{zou2020deep}. This database comprises accelerometer and gyroscope data acquired using Samsung, Xiaomi, and Huawei smartphones in unconstrained scenarios. The sampling frequency of the accelerometer and gyroscope sensors is 50 Hz. A total of 118 subjects participated in the acquisition, and both walking and non-walking sessions were considered. 

Regarding the experimental protocol of the whuGAIT database, Zou \textit{et al.} proposed in \cite{zou2020deep} a predefined division of the database into development and evaluation datasets in order to facilitate the comparison among approaches. For each subject, 90\% of the samples are considered for development while the remaining 10\% for the final evaluation. In total 33,104 samples are considered for the development dataset whereas the remaining 3,740 samples are used for the final evaluation. 

%The database is divided in 8 subsets, being the subset \#3 the one used in this study. It comprises data of all 118 users in both development and evaluation datasets. Following the experimental protocol presented by Tran \textit{et al.} in \cite{tran2021multi}, time sequences of 80 gait signals are used. Being the sampling rate 50 hz, some important information can be missing. For this reason, an overlapping of 97\% in the development set is implemented, while the testing set remains without overlapping. From the development dataset, 1500 random samples are used for training, while the remaining samples are included in the validation set. The entire evaluation dataset is considered for testing. 

\subsection{OU-ISIR database}
The OU-ISIR database was presented in \cite{Ngo2019OUISIR}. This database comprises 745 subjects, being the largest public gait biometric database to date. Data from accelerometer and gyroscope sensors were collected using three IMUs and a smartphone Motorola ME860 around the waist of the subject. The sampling frequency of the sensors is 100 Hz. Subjects had to perform 4 different activities (two flat walking, slope-up walking, and slope-down walking). The database is divided into two different subsets. The first one includes data from 744 users collected by one IMU located in the middle of the subject's back waist. The second one contains data from 408 subjects collected by the three IMUs and the smartphone.

Regarding the experimental protocol of the OU-ISIR database, we consider the predefined division of the database into development and evaluation datasets proposed by Zou \textit{et al.} in \cite{zou2020deep}. For each subject, 87.5\% of the samples are considered for development while the remaining 12.5\% for the final evaluation. In total 13,212 samples are considered for the development dataset whereas the remaining 1,409 samples are used for the final evaluation.

%The popular OU-ISIR \cite{Ngo2019OUISIR} contains two subsets. Only the first one is used in this study. A total of 745 subjects data collected by a inertial measurement unit located in the subject waist is used. Following the experimental protocol presented by Zou \textit{et al.} in \cite{zou2020deep}, the data of each subject is divided in two datasets, development and evaluation. Each time sequence contains 128 gait signals. An overlapping of 61\% in both development  and evaluation sets is implemented, The development dataset is split in training (90\% of the samples) and validation (the remaining part). The evaluation dataset is considered for testing. 

%% file: SystemConfiguration.tex
This section provides the system configuration details of the Transformers and traditional DL architectures (i.e., CNNs and RNNs) considered in the experimental framework of the study. 

Regarding the input of the models, we consider in all of them the same approach. For the whuGAIT database, a total of 80 time signals (around 1.5 seconds each) are extracted from the 3-axis accelerometer and gyroscope sensors following the approach presented in~\cite{tran2021multi}. Also, we consider an overlapping of 97\% between samples in training. For the OU-ISIR database, 128 time signals (around 1.5 seconds each) are extracted from the 3-axis accelerometer and gyroscope sensors following the approach presented in~\cite{zou2020deep}. Also, we consider an overlapping of 61\% between samples in training.

For a better comparison of Transformer architectures with popular DL architectures, we consider the following approaches: \textit{i)} CNNs, \textit{ii)} RNNs, and \textit{iii)} a hybrid configuration based on the combination of CNNs and RNNs. These DL models are widely considered for gait biometric recognition, achieving state-of-the-art results as described in Sec.~\ref{sec:relatedWork}. CNNs have shown advantages in capturing spatial dependencies, while RNNs are better to capture the temporal dependencies.

We provide next a description of the networks parameters:

\begin{itemize}
    \item \textit{CNN}: we consider four 1D convolutional layers with 6 units each and kernel size 5, followed by one dense layer with $\frac{3}{2}L$ units (where $L$ is the length of the time sequence), and one softmax layer. After every 2 convolutional layers, we use max-pooling and dropout with a 0.5 rate. ReLU activation functions are used in both convolutional and dense layers. 
    \item \textit{RNN}: we consider three LSTM layers with 3 units each followed by one dense layer with $\frac{3}{2}L$ units, and one softmax layer. 
    \item \textit{CNN-RNN}: it comprises two parallel modules, \textit{i)} four convolutional layers with 6 units each and kernel size 5, and \textit{ii)} three LSTM layers with 3 units each. After both modules, a feature concatenation is applied, followed by one dense layer with $\frac{3}{2}L$ units, and one softmax layer. We also consider dropout with 0.5 rate after each convolutional layer.
    \item \textit{Vanilla Transformer \cite{vaswani2017attention}}: we consider the positional encoding together with the encoder part of the Vanilla Transformer. The model consists of $N = 5$ layers. Regarding the multi-head Self-Attention sub-layer, 8 heads are considered with Full-Attention whereas for the point-wise feed-forward network we consider two linear layers (layer 1 with $L$ units and layer 2 with $L*4$ units) with ReLU activation and dropout in between. %While the first linear layer has $L$ units, the second linear layer has $4*L$, where $L$ is the length of each time sequence (80 for whuGAIT database and 128 for OU-ISIR database).
    \item \textit{Informer \cite{zhou2021informer}}: we consider the same structure as the Vanilla Transformer but changing in the multi-head Self-Attention sub-layer the Full-Attention to ProbSparse-Attention. The model is composed of $N = 5$ layers. Regarding the multi-head Self-Attention sub-layer, 8 heads are considered whereas for the point-wise feed-forward network we consider two linear layers (layer 1 with $L$ units and layer 2 with $L*4$ units) with ReLU activation and dropout in between.
    \item \textit{Autoformer \cite{xu2021autoformer}}: the same structure as the Vanilla Transformer is considered but changing the Self-Attention mechanism for the Auto-Correlation mechanism. The model comprises $N = 5$ layers with 8 heads in the multi-head Auto-Correlation sub-layer. For the point-wise feed-forward network we consider two linear layers (layer 1 with $L$ units and layer 2 with $L*4$ units) with ReLU activation and dropout in between.
    \item \textit{Block-Recurrent Transformer \cite{hutchins2022block}}: it comprises 12 layers: $N = 9$ multi-head Self-Attention layers with Cross-Attention and Full-Attention (8 heads), followed by $R = 1$ recurrent layer, and $N = 2$ more multi-head Self-Attention layers with Cross-Attention and Full-Attention (8 heads). In each layer, the point-wise feed-forward network is composed of two linear layers (layer 1 with $L$ units and layer 2 with $L*4$ units) with ReLU activation and dropout in between.
    \item \textit{THAT \cite{li2021two}}: this is a two-stream convolution Transformer architecture. In the first stream (Temporal Module) the time-over-channel features are analysed. To this aim, Gaussian range encoding is used together with the original multi-head Self-Attention sub-layer (Full-Attention with 8 heads). The HAR CNN sub-layer is based on a multi-scale CNN (3 convolutional layers with $L$ units each, ReLU activation functions, and kernel sizes 1, 3, and 5 respectively, followed by dropout layers). The Temporal Module contains $N = 9$ layers. Regarding the second stream (Channel Module) the data is transposed to extract the channel-over-time features, adopting the original Vanilla Transformer structure with positional encoding. The multi-head Self-Attention sub-layer contains Full-Attention with 6 heads. The HAR CNN sub-layer is based on a multi-scale CNN (3 convolutional layers with $L$ units each, ReLU activation functions, and kernel sizes 1, 3, and 5 respectively, followed by dropout layers). The Channel Module contains $H = 1$ layer.
    \item \textit{Proposed Transformer}: we consider a two-stream Transformer based on Temporal and Channel Modules. Both modules use Gaussian range encoding. Regarding the Temporal Module, it comprises 12 layers: $N = 9$ multi-head Auto-Correlation layers (8 heads), followed by $R = 1$ recurrent layer (8 heads), and $N = 2$ multi-head Auto-Correlation layers (8 heads). In each layer, the GBR CNN sub-layer is based on a multi-scale CNN (4 convolutional layers with $L$ units each, ReLU activation functions, and kernel sizes 1, 3, 5, and 7 respectively, followed by dropout layers). Regarding the Channel Module, it comprises $H = 1$ layers. In all of them we consider multi-head Auto-Correlation mechanism with 6 heads. The GBR CNN sub-layer is based on a multi-scale CNN (4 convolutional layers with $L$ units each, ReLU activation functions, and kernel sizes 1, 3, 5, and 7 respectively, followed by dropout layers).
\end{itemize}

For the training of the models, we use cross-entropy and Adam optimiser with default parameters (learning rate of 0.001). All models are adapted to the gait biometric recognition task. To this aim, after the models we include 2 convolutional layers ($L$ units each, ReLU activation functions, and kernel sizes 128, followed by dropout layers) with max-pooling and a linear layer with softmax activation function. For the THAT and proposed Transformer, we also consider feature concatenation of the Temporal and Channel Modules as described in Fig \ref{fig:TransformerArchs} \textit{E.} and \textit{F.} 

%% file: ExperimentalResults.tex
This section aims to analyse the performance of the different state-of-the-art Transformer architectures considered in this study (i.e., Vanilla, Informer, Autoformer, Block-Recurrent Transformer, THAT, and the proposed one) for the topic of gait biometric recognition on mobile devices. Sec.~\ref{comparison_DL} provides a comparison of Transformer architectures with traditional DL architectures such as CNNs and RNNs. Finally, Sec.~\ref{ComparisonSOTA} provides a comparison of the proposed Transformer architectures with the state of the art. 

\begin{table}[tp!]
\centering
\caption{Comparison of traditional DL models and recent Transformers for gait biometric recognition.}
\resizebox{\columnwidth}{!}{
\begin{tabular}{|c|c|c|}
\hline
\textbf{Database}         & \textbf{Model}       & \textbf{Accuracy} \\ \hline \hline
 
                        & CNN                  &             75.31\%      \\ \cline{2-3} 
                         & RNN                 &   82.42\%                \\ \cline{2-3} 
                         &  CNN + RNN         &            84.54\%        \\ \cline{2-3} 
                        & Vanilla Transformer                  &           87.73\%        \\ \cline{2-3} 
whuGAIT                         & Informer                 &           89.26\%         \\ \cline{2-3} 
                         & Autoformer          &           89.44\%        \\ \cline{2-3} 
                      & Block-Recurrent Transformer  &      91.78\%             \\ \cline{2-3}
                         & THAT  &      92.99\%             \\ \cline{2-3} 

                         & \textbf{Proposed Transformer} &        \textbf{94.25\%}            \\ \hline \hline
                         & CNN                  &             32.51\%     \\ \cline{2-3} 
                         & RNN                 &     44.15\%              \\ \cline{2-3} 
                         & CNN + RNN          &     46.63\%             \\ \cline{2-3} 
                        & Vanilla Transformer                   &       54.51\%        \\ \cline{2-3} 
OU-ISIR                      & Informer                 &        59.40\%           \\ \cline{2-3} 
                         & Autoformer           &         63.10\%           \\ \cline{2-3} 
                       & Block-Recurrent Transformer  &    69.98\%                \\ \cline{2-3} 
                         &  THAT  &       85.74\%           \\ \cline{2-3} 

                         & \textbf{Proposed Transformer} &           \textbf{93.33\%}         \\ \hline
\end{tabular}
}
\label{table:TransformersComparison}
\end{table}

\subsection{Transformers vs. traditional DL architectures}\label{comparison_DL}

Table~\ref{table:TransformersComparison} provides a comparison of traditional DL models and recent Transformers for the whuGAIT and OU-ISIR databases. The best results achieved for each database are remarked in bold. First, we can see that the Vanilla Transformer outperforms the traditional DL models (CNN, RNN, and CNN + RNN) in both databases. The Vanilla Transformer achieves an accuracy of 87.73\% in the whuGAIT database (absolute improvement of 3.19\% accuracy compared with the CNN + RNN approach), and 54.51\% in the OU-ISIR database (absolute improvement of 7.88\% accuracy compared with the CNN + RNN approach). These performance improvements prove the advantages of Transformers compared with traditional CNN and RNN architectures, for example, the ability to train the model using large time sequences, attending to all the previous samples at the same time. In addition, we can also observe a considerable gap in the results between the whuGAIT and OU-ISIR databases. This is produced due to the OU-ISIR is a more challenging database including many more subjects, sensors, and walking styles. This trend is also observed in the original article for traditional CNN and RNN architectures~\cite{Ngo2019OUISIR}.

The Vanilla Transformer architecture was improved using ProbSparse-Attention (Informer) and Auto-Correlation (Autoformer). Analysing the results included in Table~\ref{table:TransformersComparison}, we can observe that both Informer and Autoformer outperform the Vanilla Transformer in both whuGAIT and OU-ISIR databases. In particular, for the whuGAIT database, the Informer and Autoformer achieve 89.26\% and 89.44\% accuracy, respectively, in comparison with the 87.73\% accuracy achieved for the Vanilla Transformer (absolute improvement of around 2\% accuracy). Regarding the OU-ISIR database, much better results are achieved by Informer and Autoformer compared with the Vanilla Transformer (59.40\%, 63.10\%, and 54.51\% accuracy, respectively). Also, the Autoformer outperforms the Informer in both databases, proving the potential of the multi-head Auto-Correlation mechanism, replacing the point-wise connections for series-wise connections.

The Block-Recurrent Transformer was presented as an alternative to use the dot-product or periodicity-based series mechanism, which fix an attention window size. Analysing the results of Table~\ref{table:TransformersComparison}, the Block-Recurrent Transformer outperforms previous Transformers for both whuGAIT (91.78\% accuracy) and OU-ISIR (69.98\% accuracy) databases. This improvement is specially relevant for the OU-ISIR database with an absolute improvement of 6.88\% accuracy compared with the Autoformer.

The THAT Transformer proposed a two-stream approach based on Temporal and Channel Modules. This Transformer architecture outperforms all previous Transformers, achieving accuracies of 92.99\% and 85.74\% for the whuGAIT and OU-ISIR databases, respectively. The improvement is much higher for the OU-ISIR database with an absolute improvement of 15.76\% accuracy compared with the Block-Recurrent Transformer. The main reason for this improvement is the proposed Gaussian range encoding in the Temporal Module, capturing better the temporal information of the sample than the position encoding considered in all previous Transformers. Moreover, by having multi-scale convolutions instead of feed-forward linear layers, more discriminative patterns of each user are captured. THAT also demonstrates how, by obtaining features from two points of view (time-over-channel features and channel-over-time features), complementary information can be captured, achieving better performance.

Finally, we analyse the results achieved by our proposed Transformer architecture. As can be seen in Table~\ref{table:TransformersComparison}, the proposed Transformer outperforms all previous Transformer architectures for both whuGAIT (94.25\% accuracy) and OU-ISIR (93.33\% accuracy) databases. In particular, the proposed Transformer achieves absolute improvements of 1.26\% (THAT), 2.47\% (Block-Recurrent Transformer), 4.81\% (Autoformer), 4.99\% (Informer), and 6.52\% (Vanilla Transformer) accuracy for the whuGAIT database. This improvement is even higher for the challenging OU-ISIR database with absolute improvements of 7.59\% (THAT), 23.35\% (Block-Recurrent Transformer), 30.23\% (Autoformer), 33.93\% (Informer), and 38.82\% (Vanilla Transformer) accuracy. The improvement achieved by the proposed Transformer is produced for several reasons. First, the Gaussian range encoding allows to introduce in each sample details about its relative position with respect to the contiguous samples (before the Temporal Module) and about the different channels (before the Channel Module), obtaining more complex information. Another advantage is the two-stream architecture, where each of the modules extracts different features (the Temporal Module extracts time features while the Channel Module extracts spatial features). By extracting features from two different perspectives, a more global view of each sample is obtained. In addition, the application of Auto-Correlation in the multi-head Self-Attention mechanism together with the Gaussian range encoding in both Temporal and Channel Modules allow to extract series-wise connections in each range of the encoding, analysing the different behaviour of each sample in different environments. Furthermore, including the recurrent layer proposed in the Block-Recurrent Transformer to the Temporal Module offers a more complete analysis. The module summarises all the information seen previously, giving a more global view of each sample with respect to the rest. Finally, by including a multi-scale CNN instead of the original feed-forward network, the whole model is series-wise: from the Gaussian range encoding that extracts the position of each sample based on a range of points, multi-head Auto-Correlation with Block-Recurrent Attention, which extracts information periodically based on series, and multi-scale CNN that applies convolutions with different kernels to test the behaviour of samples in different ranges.

\begin{figure}[tp!]
\centering
\includegraphics[width=\linewidth]{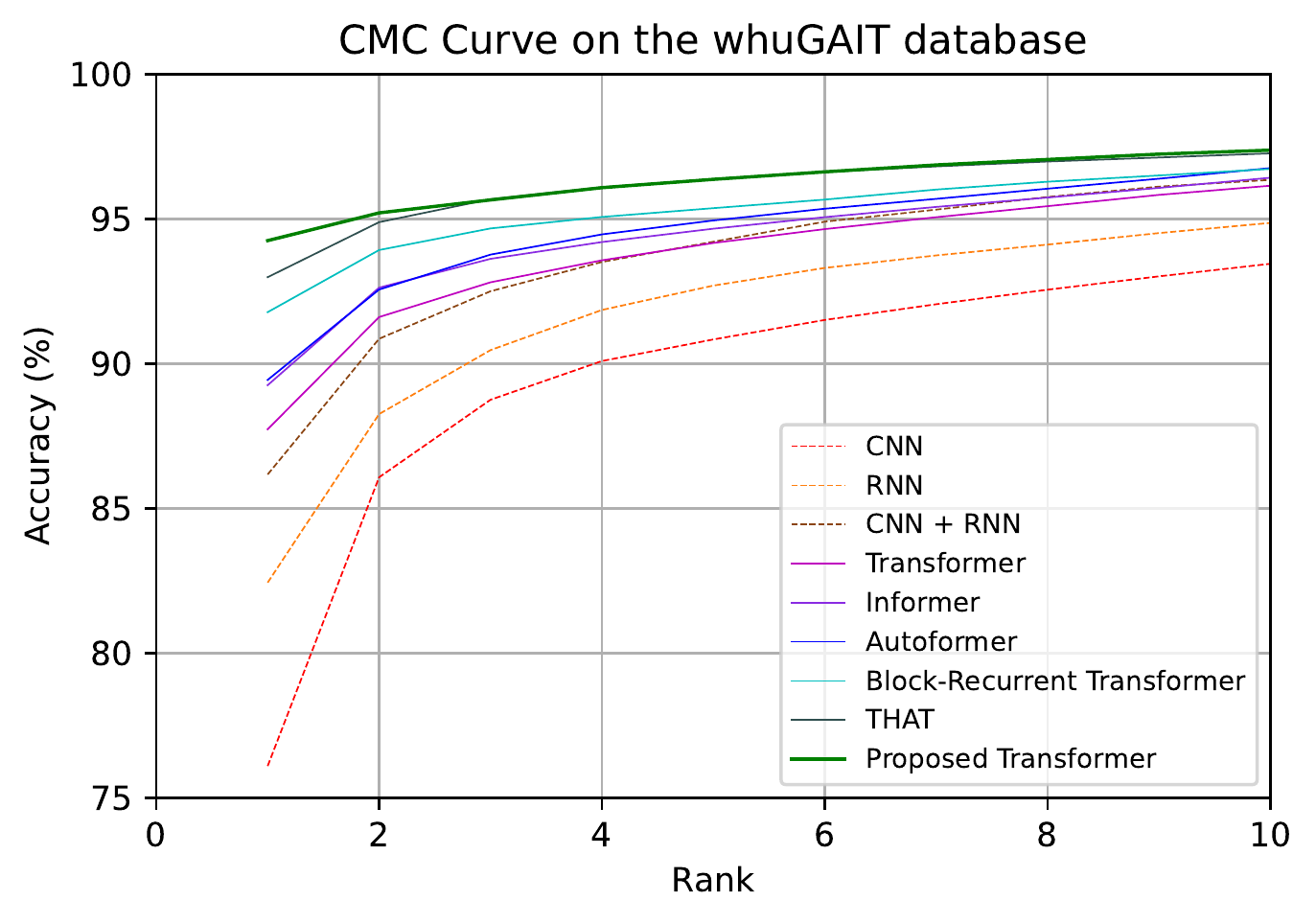}
\hfill
\includegraphics[width=\linewidth]{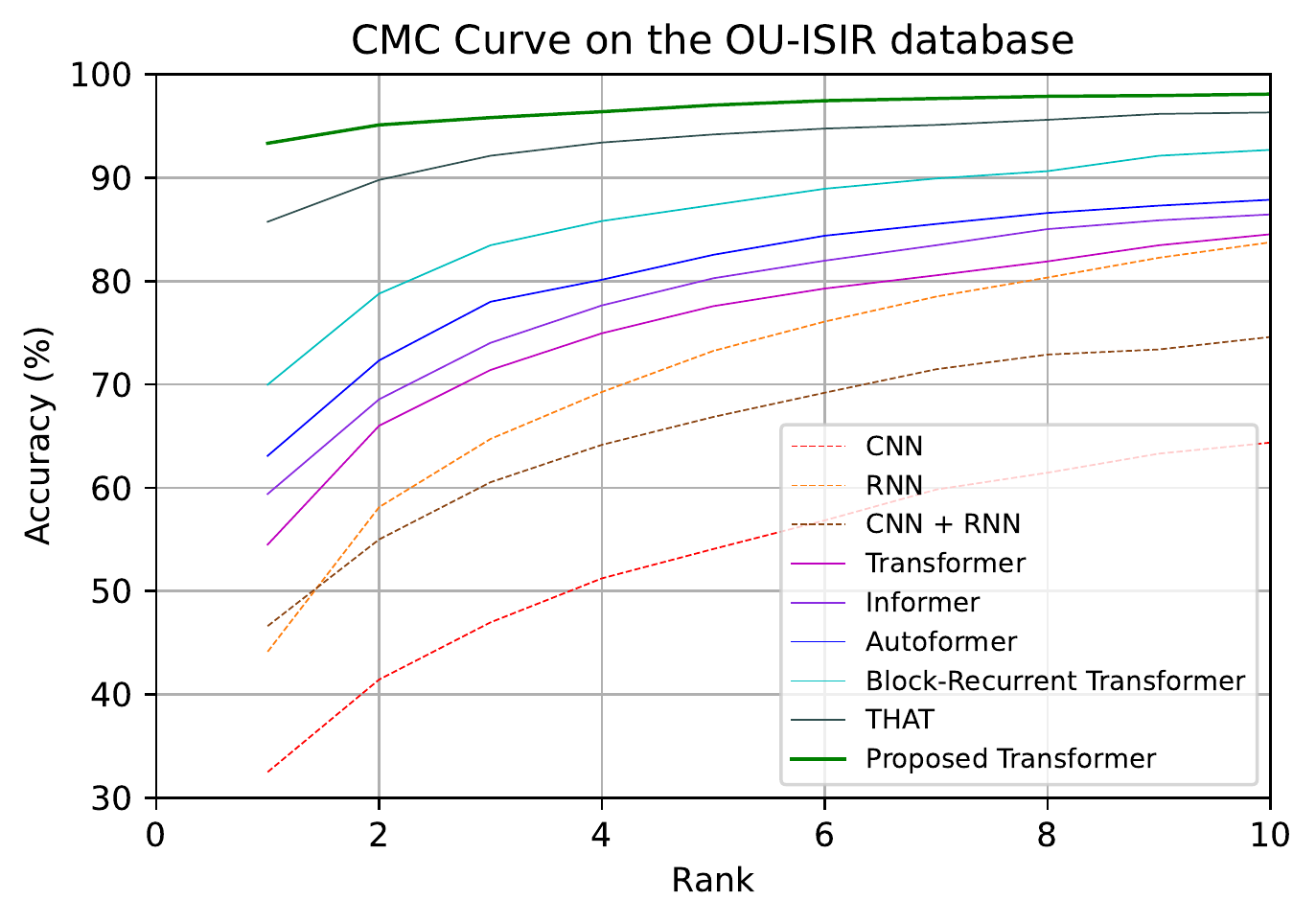}
\caption{Cumulative Match Characteristic (CMC) curves of the traditional DL models (CNN, RNN, CNN + RNN) and recent Transformers (Vanilla, Informer, Autoformer, Block-Recurrent, THAT, and the proposed Transformer) for both whuGAIT (top) and OU-ISIR (bottom) databases.}
\label{fig:CMC}
\end{figure}

Previous results correspond to the Rank-1 accuracy. Nevertheless, in some applications we might be interested in having a ranked list of possible subjects of interest (e.g., in forensic applications). Fig. \ref{fig:CMC} shows the Cumulative Match Characteristic (CMC) curve of the traditional DL models commonly used in biometric recognition (CNN, RNN, CNN + RNN) and recent Transformers (Vanilla, Informer, Autoformer, Block-Recurrent, THAT, and the proposed Transformer) for both whuGAIT and OU-ISIR databases. In general, we can see the same trend in both databases for all approaches, improving the accuracy results with the Rank values. For example, for the proposed Transformer, the accuracy increases from 94.25\% (Rank-1) to 97.37\% (Rank-10) for the whuGAIT database whereas for the OU-ISIR database this value increases from 93.33\% (Rank-1) to 98.08\% (Rank-10).

\subsection{Comparison with the State of the Art}\label{ComparisonSOTA}

Finally, we compare in Table \ref{table:stateoftheartComparison} the Rank-1 accuracy results achieved by our proposed Transformer with other state-of-the-art approaches presented in the literature for gait biometric recognition: CNNs + SVM~\cite{gadaleta2018idnet}, RNNs~\cite{zou2020deep,tran2021multi}, and CNNs + RNNs~\cite{zou2020deep,tran2021multi,ordonez2016deep}. The best results achieved for each database are remarked in bold. It is important to highlight that all studies consider the same experimental protocol~\cite{zou2020deep} for both whuGAIT and OU-ISIR databases. 

In general, the proposed Transformer has outperformed previous approaches in both databases. For the whuGAIT database, the proposed Transformer achieves 94.25\% accuracy, showing slightly better results compared with the CNNs + RNNs approach presented in~\cite{tran2021multi}. Analysing the OU-ISIR database, the proposed Transformer further improves the results achieved by previous approaches with 93.33\% accuracy. This is an absolute improvement of 3.54\% accuracy compared with the best previous approach (CNNs + RNNs~\cite{tran2021multi}). The authors improved the CNN + RNN architecture using an RNN to process each channel, combined in parallel with a CNN with two channels, one for each sensor. These results support the high potential of the proposed Transformer for gait biometric recognition. In addition, it is important to highlight the better time complexity and memory usage of the proposed Transformer compared with traditional DL models.

\begin{table}[tp!]
\centering
\caption{Comparison of the proposed Transformer with state-of-the-art gait biometric recognition approaches.}
\resizebox{\columnwidth}{!}{
\begin{tabular}{|c|c|c|}
\hline
\textbf{Database}         & \textbf{Model}                                              & \textbf{Accuracy}            \\ \hline
                    & CNN + SVM\cite{gadaleta2018idnet}              & 92.91\% 
                    \\ \cline{2-3}
                    & RNN  \cite{zou2020deep}            & 91.88\%                      \\ \cline{2-3}      
                     & RNN  \cite{tran2021multi}            & 93.14\%                      \\ \cline{2-3}
       whuGAIT               & CNN + RNN\cite{ordonez2016deep}              & 92.25\%                      \\ \cline{2-3} 
                     & CNN + RNN  \cite{zou2020deep}            & 93.52\%                      \\ \cline{2-3}

                         & CNN + RNN \cite{tran2021multi}            & 94.15\%                      \\ \cline{2-3} 
                         %& CNN + CEDS \cite{huang2021lightweight}     & 94.71\%                      \\ \cline{2-3} 
%                         & Lightweight CNN \cite{hasan2022gait}       & 95.16\%                      \\ \cline{2-3} 
%                         & Lightweight CNN + ARB \cite{hasan2022gait} & 95.35\% \\ \cline{2-3} 
                         & \textbf{Proposed Transformer}                         &   \textbf{94.25\%}                                         \\ \hline \hline
                         
                                                     & CNN + SVM\cite{gadaleta2018idnet}             & 44.29\%                     
                                            \\ \cline{2-3}
                         & RNN  \cite{tran2021multi}            & 78.92\%                      \\ \cline{2-3}                   
  OU-ISIR                        & CNN + RNN  \cite{ordonez2016deep}              & 37.33\%   \\ \cline{2-3}

                         & CNN + RNN \cite{tran2021multi}            & 89.79\%                      \\ \cline{2-3} 
                         %& CNN + CEDS \cite{huang2021lightweight}     & 97.16\%                      \\ \cline{2-3} 
%                         & Lightweight CNN \cite{hasan2022gait}       & 98.39\% \\ \cline{2-3} 
%                         & Lightweight CNN + ARB \cite{hasan2022gait} & 98.86\%                      \\ \cline{2-3} 
                         & \textbf{Proposed Transformer}                            &           \textbf{93.33\%}                              \\ \hline
\end{tabular}
\label{table:stateoftheartComparison}
}
\end{table}

%Furthermore, Huang \textit{et al.} \cite{huang2021lightweight} introduced a lightweight Attention-based CNN improving the results of the previous works, achieving 94.71\% in the whuGAIT database, and 97.16\% in the OU-ISIR database.

%% file: Conclusions.tex
This article has explored and proposed novel behavioural biometric systems based on Transformers. To the best of our knowledge, this is the first study that presents a complete framework for the use of Transformers in gait biometrics. Several state-of-the-art Transformer architectures (Vanilla, Informer, Autoformer, Block-Recurrent Transformer, and THAT) are considered in the experimental framework, together with a new proposed configuration. Two popular public databases are considered in the analysis, whuGAIT and OU-ISIR. 

The proposed Transformer has outperformed previous Transformer architectures and traditional DL architectures (i.e., CNNs, RNNs, and CNNs + RNNs) when evaluated using both databases. In particular, for the challenging OU-ISIR database, the proposed Transformer achieves 93.33\% accuracy, resulting in accuracy improvements compared with other techniques of 7.59\% (THAT), 23.35\% (Block-Recurrent Transformer), 30.23\% (Autoformer), 33.93\% (Informer), 38.82\% (Vanilla Transformer). The proposed Transformer has also been compared with state-of-the-art gait biometric recognition systems, outperforming the results presented in the literature. In addition, it is important to highlight the better time complexity and memory usage of the proposed Transformer compared with traditional DL models.

Future work will be oriented towards analysing the potential of the proposed Transformer architecture for other behavioural biometric modalities such as handwritten signature~\cite{2021_Arxiv_SVConGoing_Tolosana}, electrocardiograms~\cite{melzi2022ecg}, and keystroke~\cite{mondal2017person}, and also its possible application for the reduction of sensitive data in biometric scenarios~\cite{delgado2021gaitprivacyon}.